\documentclass[letterpaper]{article} 
\usepackage[submission]{aaai2026} 
\usepackage{times} 
\usepackage{helvet} 
\usepackage{courier} 
\usepackage[hyphens]{url} 
\usepackage{graphicx} 
\usepackage{natbib} 
\usepackage{caption} 
\usepackage{algorithm}
\usepackage{algorithmic}
\usepackage{pifont}
\usepackage[switch]{lineno}

\usepackage{amsmath}
\usepackage[capitalise]{cleveref}
\usepackage{multirow}
\usepackage{booktabs}
\usepackage{amsfonts}
\usepackage{colortbl}
\usepackage{makecell}
\usepackage{microtype}
\usepackage{tikz}
\usepackage{wasysym}

\definecolor{light-gray}{gray}{0.9}

\usepackage{newfloat}
\usepackage{listings}
\DeclareCaptionStyle{ruled}{labelfont=normalfont,labelsep=colon,strut=off} 
\floatstyle{ruled}
\newfloat{listing}{tb}{lst}{}
\floatname{listing}{Listing}
\title{Beyond the Horizon: Decoupling Multi-View UAV Action Recognition via Partial Order Transfer}

\author{
Wenxuan Liu$^{1,2}$, Zhuo Zhou$^{3}$, Xuemei Jia$^{3}$, Siyuan Yang$^{4}$, Wenxin Huang$^{5}$, \\ 
Xian Zhong$^{2}$\footnote{Corresponding author.}, Chia-Wen Lin$^{6}$
}
\affiliations{
$^1$Peking University \quad $^2$Wuhan University of Technology \quad $^3$Wuhan University \quad $^4$Nanyang Technological University \\
$^5$Hubei University \quad $^6$National Tsing Hua University \\

}
\usepackage{bibentry}

\begin{document}

\maketitle

\begin{abstract}

Action recognition in unmanned aerial vehicles (UAVs) poses unique challenges due to significant view variations along the vertical spatial axis. Unlike traditional ground-based settings, UAVs capture actions at a wide range of altitudes, resulting in considerable appearance discrepancies. We introduce a multi-view formulation tailored to varying UAV altitudes and empirically observe a \textbf{partial order} among views, where recognition accuracy consistently decreases as altitude increases. This observation motivates a novel approach that explicitly models the hierarchical structure of UAV views to improve recognition performance across altitudes. To this end, we propose the Partial Order Guided Multi-View Network (POG-MVNet), designed to address drastic view variations by effectively leveraging view-dependent information across different altitude levels. The framework comprises three key components: a \textbf{View Partition (VP)} module, which uses the head-to-body ratio to group views by altitude; an \textbf{Order-aware Feature Decoupling (OFD)} module, which disentangles action-relevant and view-specific features under partial order guidance; and an \textbf{Action Partial Order Guide (APOG)}, which uses the partial order to transfer informative knowledge from easier views to more challenging ones. We conduct experiments on \textsc{Drone-Action}, \textsc{MOD20}, and \textsc{UAV}, demonstrating that POG-MVNet significantly outperforms competing methods. For example, POG-MVNet achieves a 4.7\% improvement on \textsc{Drone-Action} and a 3.5\% improvement on \textsc{UAV} compared to state-of-the-art methods ASAT and FAR. Code will be released soon.

\end{abstract}


\section{Introduction}

Human action recognition is widely studied~\cite{Feichtenhofer0M19, Feichtenhofer20, ZhouLXWZ23, YangLLEHK24}, primarily based on videos captured by ground cameras~\cite{liuwenxuan}. With the rapid development of unmanned aerial vehicles (UAVs), new opportunities have emerged for aerial action recognition in various fields.
Early works~\cite{PereraLC18, perera2019drone, LiLZNWL21} explore the use of pose estimation or attention mechanisms~\cite{Cheng0HC0L20} to address view changes. However, they overlook a crucial aspect: UAVs operate in an open aerial space with high mobility. The fast movement of UAVs causes drastic shifts in view angle and distance, altering the visual appearance of actions. Ignoring these variations often \textbf{biases models toward low-altitude views}, impairing recognition at higher altitudes.

\begin{figure}[t]
	\centering
	\includegraphics[width = \linewidth]{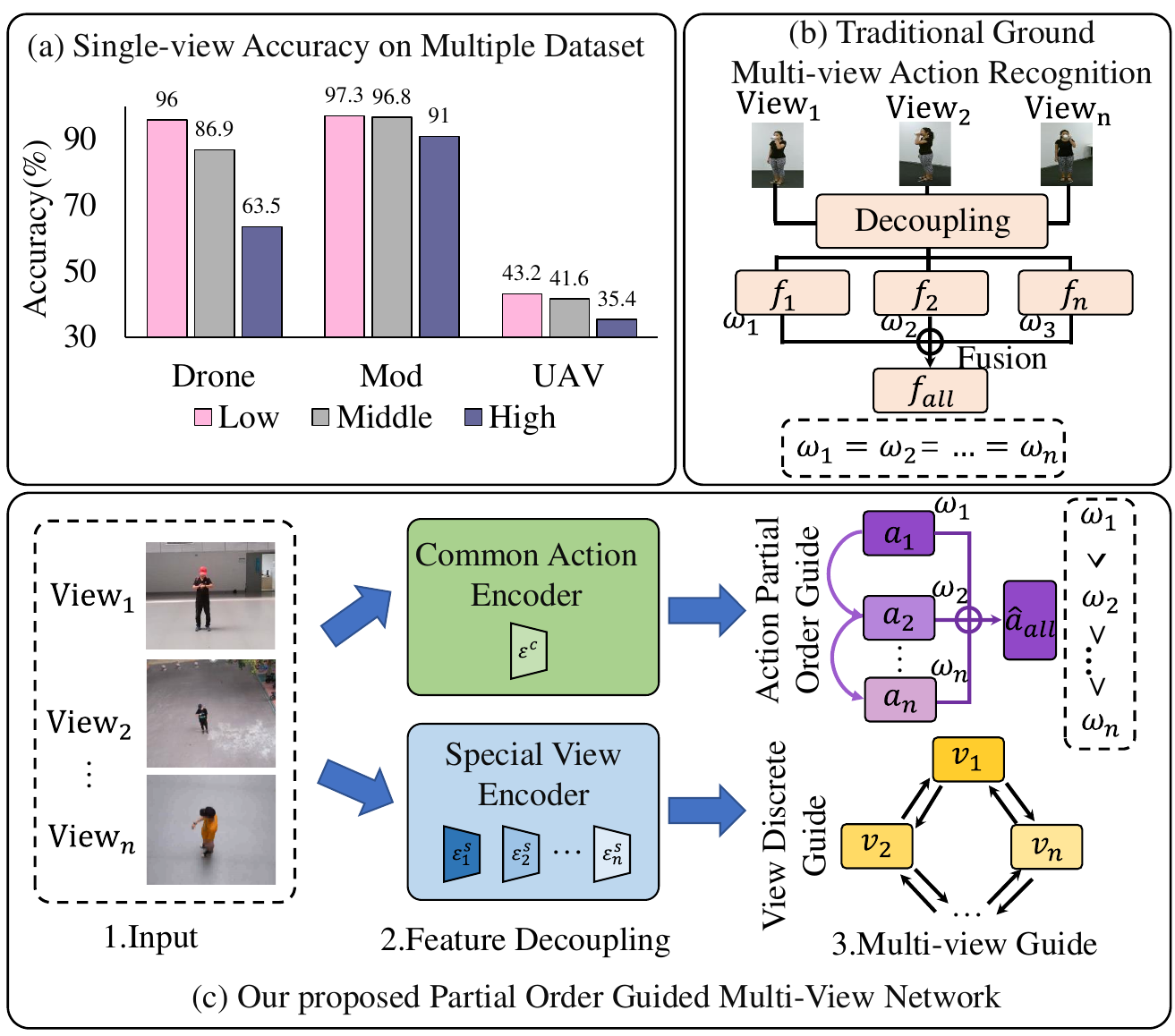} 
	\caption{(a) Recognition accuracy per view on \textsc{Drone-Action}, \textsc{MOD20}, and \textsc{UAV}, highlighting discrepancies across individual views. (b) Traditional ground-based multi-view recognition pipeline. (c) Our OFD module, comprising a feature decoupling unit and two guidance units: action guide and view guide.}
	\label{fig-motivation}
\end{figure}

Building on this, we revisit view variation in UAV action recognition and observe a systematic pattern along the vertical axis, unlike the arbitrary changes in ground-based settings~\cite{LiuZZJWL23}. We find that as viewing altitude increases, actions become increasingly ambiguous and harder to recognize due to greater visual variability. To quantify this, we use the head-to-body ratio as an interpretable proxy for altitude. As shown in \cref{fig-motivation}(a), recognition accuracy consistently declines as this ratio increases, revealing a \textbf{partial order} among UAV views.

Motivated by this observation, we propose a novel framework, the Partial Order Guided Multi-View Network (POG-MVNet), which fully exploits these partial orders and addresses two core questions:

\textbf{\textit{Q1: How can we isolate action-relevant features from view-induced variations under the partial order?}} While traditional multi-view action recognition methods~\cite{UllahMHB21, XuWLJWL21, ZhongZLJJHW22, LiuZZJWL23} explore view-invariant modeling or feature disentanglement, they assume all views contribute equally to recognition. As shown in \cref{fig-motivation}(b), average feature fusion implicitly neglects the structured disparity across views. In contrast, we propose an \textbf{Order-aware Feature Decoupling (OFD)} module that explicitly captures action-relevant features in a view-sensitive manner. Specifically, OFD disentangles each input into view-invariant action features and view-specific components via a shared encoder and multiple view branches. A generative loss further encourages clear separation of action and view information, yielding more robust and transferable action representations. Furthermore, rather than treating all views uniformly, we model their hierarchical structure based on the observed partial order, using it to guide both the decoupling and integration of features across views.



\textbf{\textit{Q2: How can we exploit the partial order among views to guide adaptive knowledge transfer from easier to harder views?}} Building on the disentangled features from OFD, we introduce an \textbf{Action Partial Order Guide (APOG)} unit to facilitate structured knowledge transfer across views. APOG aligns action features according to their position in the partial order, enabling progressive adaptation from low- to high-altitude views. We construct a graph over view features to discretize the continuous view space and amplify distribution gaps between views. This design reduces residual coupling of view and action features and ensures transfer follows the intrinsic view hierarchy. By integrating OFD with APOG, POG-MVNet fully leverages the partial order among UAV views, leading to more effective and robust action recognition under diverse viewing conditions.

In summary, the contributions of this work are threefold:

\begin{itemize}
	\item We introduce a multi-view formulation for UAV action recognition, adopting the head-to-body ratio for view partitioning and establishing a novel classification foundation.

	\item We reveal a partial order among UAV multi-view settings, addressing vertical spatial view variation, and propose an Order-aware Feature Decoupling (OFD) module to tackle multi-view challenges.
	
	\item We propose Partial Order Guided Multi-View Network (POG-MVNet) to guide the learning and integration of UAV multi-view information via partial order relations, achieving significant improvements over state-of-the-art methods.

\end{itemize}

\section{Related Work}

\subsection{Multi-View Action Recognition}

\subsubsection{UAV View.} 

Early research~\cite{PereraLC18, perera2019drone, LiLZNWL21} applies pose estimation~\cite{Cheng0HC0L20} to extract local action features from the global receptive field in UAV videos. However, conventional feature extractors struggle with small targets, resulting in noisy representations. Subsequent works~\cite{JinMHXZ22, KothandaramanGW22} shift focus to temporal modeling, emphasizing dynamic motion characteristics. More recent studies~\cite{Xian_2024_WACV, Xian_2024_WACV1} explore sampling strategies via feature-based representations, while \cite{Peng_Shu_Yao_Xie_2025} introduce a token-level compression mechanism built on a transformer block.


\subsubsection{Ground Multi-View.} 
Recent studies on cross-view action recognition aim to learn view-invariant representations~\cite{UllahMHB21, XuWLJWL21, LiuZZJWL23, YangZXZC023, You2024ConvertingAN}. Virtual camera simulation~\cite{RahmaniMS18, XiaoCWCZB19} facilitates multi-view learning but is annotation-heavy and computationally expensive. Alternatives based on feature clustering~\cite{ShaoLZ21, UllahMHB21} often lack strong constraints, limiting generalization to unseen views. Feature distillation approaches~\cite{VyasRS20, ZhongZLJJHW22, LiuZZJWL23, SiddiquiTS24}, \textit{e.g.}, CVAM for scene dynamics or VCD and DRDN for disentangling view/action features, improve robustness but overlook inter-view relations. In contrast, we explicitly decouple action and view features by leveraging the \textbf{partial order} among views to guide discriminative feature learning.


\subsection{Sample Classification} 
Differentiation of instances has been studied from various perspectives, including confidence-based learning~\cite{HanYYNXHTS18, WanXWYZH22a} and meta-learning~\cite{0001HLF021, WangJYCGG23, tmm/ZhongGYHL23}. Some methods leverage measures like prediction variance or threshold closeness to distinguish instances, while others exploit ordinal relations by recoding labels for classification and ranking~\cite{NiuZWGH16, WangJYCGG23}. In contrast, we propose a novel partitioning strategy that leverages the inherent structure of multi-view UAV data. By grouping instances based on head-to-body ratio, we enable more effective utilization of view-specific cues for action recognition.

\subsection{Transfer Learning}

Our problem centers on transferring knowledge from a baseline to a compact model. Unlike standard transfer learning methods, such as distillation~\cite{0004TLHWCJ020}, pre-trained initialization~\cite{Bolya2021ScalableDM}, or rehearsal~\cite{JiHLPL23, SunLZWG23, Tian2023KnowledgeDW}, which focus on classification and ignore structural data variation, we address view-aware action recognition. By leveraging the partial order among views and partitioning instances based on head-to-body ratio, we enable more effective feature-level knowledge transfer across perspectives.

\begin{figure*} 
	\centering
	\includegraphics[width = \linewidth]{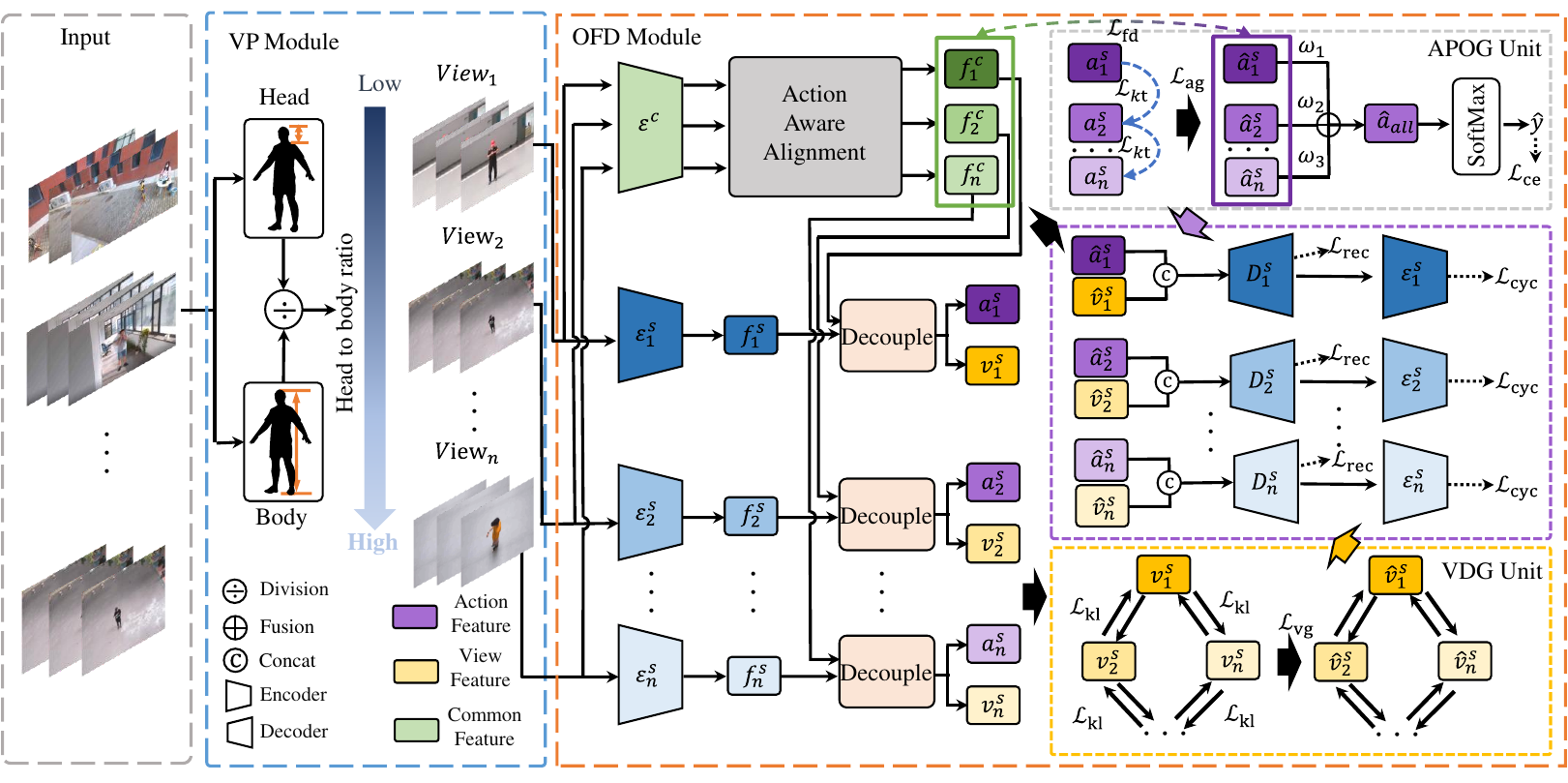} 
	\caption{\textbf{Framework of the Proposed POG-MVNet.} The VP module divides UAV samples into altitude-based views using the head-to-body ratio. The OFD module disentangles and transfers action features from low- to high-altitude views via the partial order relation and view-specific weights. The APOG unit learns this partial order to guide feature transfer, and the VDG unit imposes discretization constraints on view features to decouple multi-view representations.}
	\label{fig-framework}
\end{figure*}

\section{Proposed Method}


We propose a Partial Order Guided Multi-View Network (POG-MVNet), illustrated in \cref{fig-framework}. POG-MVNet comprises two main components: the View Partition (VP) module and the Order-aware Feature Decoupling (OFD) module. The VP module partitions samples into low-, mid-, and high-altitude views, while the OFD module decouples multi-view features and learns discriminative action representations guided by the partial order among views.

\subsubsection{Notation.} 
We adopt a classification backbone (\textit{e.g.}, X3D~\cite{Feichtenhofer20}) as a feature extractor. Let $\epsilon^c_i$ denote the common feature extractor, which maps the $i$-th view input $X_i$ to a common feature:
\begin{equation}
	f^c_i = \epsilon^c_i \left(X_i \right).
\end{equation}
We also employ view-specific extractors $\epsilon^s_i$ to obtain special features:
\begin{equation}
	f^s_i = \epsilon^s_i \left(X_i \right).
\end{equation}
The VP module computes the head-to-body ratio $H$ for each sample to guide view partition. The OFD module then decouples each $f^s_i$ into an action feature $a^s_i$ and a view feature $v^s_i$, refines $a^s_i$ according to the partial order to produce $\hat{a}^s_i$, and enhances $v^s_i$ to yield $\hat{v}^s_i$. Finally, the refined action features are fused into a global feature $\hat{a}_{\mathrm{all}}$ for classification.

\subsection{View Partition Module} \label{sec:View Partition}

To model multi-view relations, we introduce the View Partition (VP) module, which groups UAV samples into low-, mid-, and high-altitude views based on the head-to-body ratio. Specifically, we use YOLOv8\footnote{\url{https://github.com/ultralytics/ultralytics}} to detect the actor’s head and body bounding boxes, then compute the ratio:
\begin{equation}
	H = \frac{B_\mathrm{head}}{B_\mathrm{body}}, 
\label{eq1} 
\end{equation}
where $B_\mathrm{head}$ and $B_\mathrm{body}$ are the heights of the head and body bounding boxes detected by YOLOv8. A smaller $H$ indicates a larger field of view (\textit{i.e.}, lower-altitude view) and thus richer action detail. We then sort all samples by $H$ in ascending order and divide them into $n$ equal groups, assigning each sample a view index $v_i$ based on its position in this sorted list:
\begin{equation}
	v_i = \begin{cases}
	0, & 0 < \tfrac{i}{m} \le \tfrac{1}{n}, \\
	1, & \tfrac{1}{n} < \tfrac{i}{m} \le \tfrac{2}{n}, \\
	\vdots \\
	n-1, & \tfrac{n-1}{n} < \tfrac{i}{m} \le 1,
	\end{cases}
\end{equation}
where $v_i$ denotes the view index of the $i$-th sample, $n$ is the number of altitude-based views, and $m$ is the total number of samples. After sorting all samples by $H$ in ascending order, we divide them into $n$ equal groups, assigning each sample its corresponding $v_i$. This partitioning not only enables detailed multi-view analysis across altitude levels but also allows exploration of nuanced variations in head-to-body ratios throughout the dataset, providing deeper insights into the dynamics of UAV action recognition in vertical spatial contexts.

\begin{figure}
	\centering
	\includegraphics[width = 0.9\linewidth]{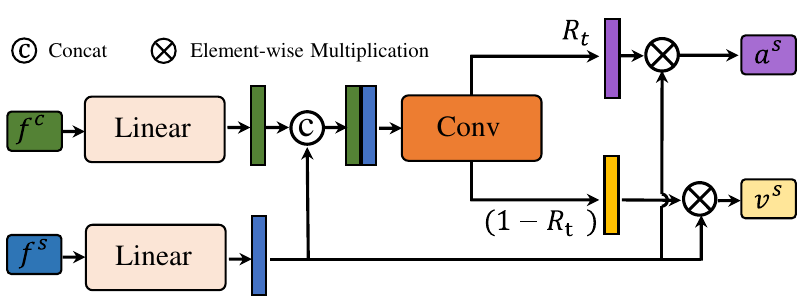} 
	\caption{\textbf{UAV Multi-View Feature Decoupling Unit.} Given a common feature $f^c$ and a view-specific feature $f^s$, we compute the action correlation map $R_t$ and decouple $f^s$ into an action feature $a^s$ and a view feature $v^s$.}
	\label{fig-guide}
\end{figure}

\subsection{Order-Aware Feature Decoupling Module}


\subsubsection{UAVs Multi-View Feature Decoupling.}



\cref{fig-guide} illustrates our decoupling process. Given the common feature $f^c_i$ and view-specific feature $f^s_i$, we first align their dimensions via a linear projection of $f^c_i$ and concatenate:
\begin{equation}
	R_t = \sigma \left(W_r \left[f^c_i, f^s_i \right] \right), 
\label{eq-rt}
\end{equation}
where $[\cdot,\cdot]$ denotes concatenation, $W_r$ comprises two successive $1\times1$ convolutional layers with batch normalization (BN) and rectified linear unit (ReLU), and $\sigma$ is the sigmoid activation. The resulting correlation map $R_t$ is then element-wise multiplied with the view-specific feature $f^s_i$ to produce the action feature $a^s_i$ and view feature $v^s_i$:
\begin{equation}
	a^s_i = f^s_i \otimes R_t, \quad
	v^s_i = f^s_i \otimes \left(1 - R_t \right), 
\label{eq-decouple}
\end{equation}
where $\otimes$ denoting element-wise multiplication, producing the action feature $a^s_i$ and view feature $v^s_i$ for view $i$.


\begin{figure}
	\centering
	\includegraphics[width = 0.9\linewidth]{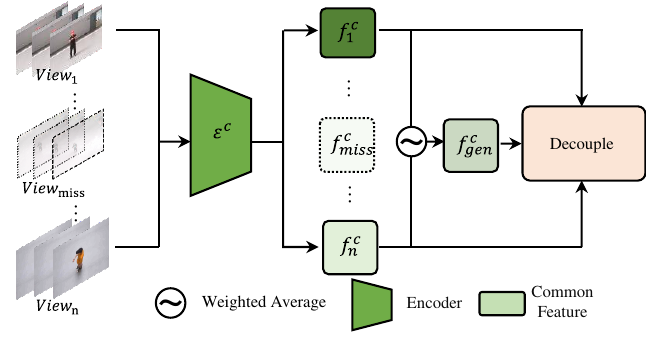} 
	\caption{\textbf{Proposed Action-Aware Alignment Unit.} The missing common feature $f^c_{\mathrm{miss}}$ is compensated by averaging the common features from the other views.}
	\label{fig-action-aware}
\end{figure}

\subsubsection{Action-Aware Alignment.} 


Unrestricted UAV views may lead to missing action categories in certain views (see \cref{fig-action-aware}). To keep visual information, if the $j$-th view input $X_j$ lacks its action representation, we complement its common feature by averaging the available features from the other views:
\begin{equation}
	f^c_j = \frac{1}{n-1} \sum_{i\neq j} f^c_i,
\label{eq-aware}
\end{equation}
where $f^c_j$ is exactly the synthesized feature $f^c_\mathrm{gen}$ for the missing view.


\subsubsection{Action Partial Order Guide.}

After the view partitioning, we compute a credibility score for each view to quantify its reliability under the partial order:
\begin{equation}
	c_i = \frac{n}{D}, \quad D = \sum_{k = 1}^n \left(d_k + 1 \right) = \sum_{k = 1}^n \alpha_k, 
\label{eq2}
\end{equation}
where $n$ is the number of views, $d_k$ is the ReLU-activated prediction score for view $k$, and $D$ represents the overall Dirichlet strength. These scores form a set of weights $\{c_i\}_{i=1}^n$ that reflect each view’s confidence. Let $A=\{a_1, a_2, \dots, a_n\}$ be the set of action features from all views, and let $a_s\in A$ be a selected source feature from an easier (low-altitude) view. We then transfer knowledge to each target feature $a_i$ by minimizing:
\begin{equation}
	\mathcal{L}_\mathrm{kt} = \frac{1}{n} \sum_{i = 1}^n c_i \lVert a_s, a_i \rVert_F^2.
\label{eq2}
\end{equation}
This adaptive, confidence-weighted transfer enables informative low-altitude views to guide learning in higher-altitude views. After transfer, we obtain a refined set of action features under the partial order, $\hat{A} = \{\hat{a}_1, \hat{a}_2, \dots, \hat{a}_n\}$, where $\hat{a}_i$ is the feature for the view ranked $i$ post-transfer. We then leverage these refined features to further optimize the common feature extractor: each $\hat{a}_i$ guides the learning of its corresponding common feature $f^c_i$ by minimizing:
\begin{equation}
	\mathcal{L}_\mathrm{fd} = \frac{1}{n} \sum_{i = 1}^n \lVert \hat{a}_i, f_i^c \rVert_F^2.
\label{eq2}
\end{equation}
This process reinforces alignment between transferred action representations and shared multi-view features. We then define the overall guide loss as the sum of the transfer and feature-alignment terms:
\begin{equation}
	\mathcal{L}_\mathrm{ag} = \mathcal{L}_\mathrm{kt} + \mathcal{L}_\mathrm{fd}.
\label{eq2}
\end{equation}



\begin{table*}[t]
	\centering
	\small
	\setlength{\tabcolsep}{1mm}
	\begin{tabular}{c|lcc|ccc|c}
	\toprule[1.1pt]
	Type & Method & Backbone & Multi-view & \textsc{Drone-Action} & \textsc{MOD20} & \textsc{UAV} & Params (M) \\
	\midrule
	\multirow{7}{*}{Vanilla}
	& SlowFast~\cite{Feichtenhofer0M19} $\dag$ & ResNet-50 & \Circle & 82.6 & 93.1 & 30.1 & 33.7 \\
	& X3D~\cite{Feichtenhofer20} $\dag$ & ResNet-50 & \Circle & 83.4 & 95.7 & 32.3 & \textbf{3.6} \\
	& 3DResNet + ATFR~\cite{FayyazRDN0GG21} $\dag$ & X3D & \Circle & 79.5 & 94.2 & 28.2 & 21.1 \\
	& TSM~\cite{LinGWH22} $\dag$ & ResNet-50 & \Circle & 75.4 & \underline{96.5} & 24.3 & 24.3 \\
	& SBP~\cite{Cheng2022StochasticBA} $\dag$ & Video Swin-T & \Circle & 81.2 & 96.3 & 29.1 & \underline{8.6} \\
	& UniFormerV2~\cite{abs-2211-09552} $\dag$ & CLIP-ViT & \Circle & 81.9 & 93.5 & 32.4 & 354.0 \\
	& AIM~\cite{YangZXZC023} $\dag$ & ViT-B & \Circle & 84.7 & 96.2 & 24.2 & 100.0 \\
	\midrule
	\multirow{2}{*}{Multi-View}
	& DRDN~\cite{LiuZZJWL23} $\dag$ & SlowFast & \CIRCLE & \underline{94.4} & 96.3 & 38.5 & 35.8 \\
	& DVANer~(Siddiqui et al. 2024) $\dag$ & 3D-CNN & \CIRCLE & 93.1 & 96.1 & 37.9 & 45.3 \\
	\midrule
	\multirow{4}{*}{UAVs}
	& FAR~\cite{KothandaramanGW22} & X3D & \Circle & 92.7 & - & 38.6 & 14.4 \\
	& ASAT~\cite{ShiFCZ23} & ResNet-50 & \Circle & - & \underline{98.2}$\ast$ & 39.7 & 61.8 \\
	& StaRNet~\cite{jitteruav} & SlowFast & \Circle & 85.4 & - & \underline{40.8} & 33.7 \\
	\rowcolor{light-gray}
	\cellcolor{white} & POG-MVNet (Ours) & X3D & \CIRCLE & \textbf{97.4} & \textbf{97.8} (\textbf{98.6}$\ast$) & \textbf{43.2} & 15.2 \\
	\bottomrule[1.1pt]
	\end{tabular}
	\caption{\textbf{Comparison of Top-1 accuracy (\%) and parameter count (M) against state-of-the-art methods on \textsc{Drone-Action}, \textsc{MOD20}, and \textsc{UAV}.} Best and second-best results are shown in bold and underlined, respectively. $\dag$ indicates reproduced results; $\ast$ denotes results on \textsc{MOD20} subset.}
	\label{table1}
\end{table*}

\subsubsection{View Discrete Guide.} 

To adaptively discretize the view space, we construct a directed graph whose nodes correspond to each view $v_i$ and whose edge weights $w(v_i, v_j)$ encode the discretization strength between views. Specifically, we define the distance between $v_i$ and $v_j$ as the difference of their logits, denoted by $\beta(v_i, v_j)$, and assemble these distances into a discretization matrix $E$ with entries $E_{ij} = \beta(v_i, v_j)$. We then encode both the view logits and their feature representations into the graph edges and normalize the weights via a softmax to obtain adaptive discretization strengths for downstream guidance:
\begin{equation}
	w \left(v_i, v_j \right) = \lVert v_i, v_j \rVert_F^2.
\label{eq11}
\end{equation}
We then construct a learnable edge-weight matrix $W$, where each weight $w(v_i, v_j)$ is initialized from the discretization distances in $E$ and subsequently refined through repeated graph updates. To mitigate scale differences, we normalize $W$ with a softmax over each row. Finally, we define the graph discretization loss across all views as:
\begin{equation}
	\mathcal{L}_\mathrm{vg} = \lVert W \otimes E \rVert, 
\label{eq11}
\end{equation}
which encourages the learned edge weights to respect the original view-to-view distances encoded in $E$, thereby learning adaptive inter-view interactions. The resulting discretization graph provides a foundation for flexible, data-driven view partitioning: by automatically adjusting discretization strengths, it captures diverse inter-view relations. Finally, we apply these learned weights to transform each view feature $v^s_i$ into its discretized counterpart $\hat{v}^s_i$, completing the view-specific refinement.


To further enforce separation of action and view information and reduce feature ambiguity, we first concatenate the refined features $[\hat{a}^s_i, \hat{v}^s_i]$ for each view and pass them through a decoder $D_i$ to reconstruct the original input:
\begin{equation}
	\mathcal{L}_\mathrm{rec} = \lVert X_i - D_i \left(\left[\hat{a}^s_i, \hat{v}^s_i \right] \right) \rVert_F^2,
\label{eq11}
\end{equation}
which ensures that the re-encoded view-specific features faithfully match the original special features:
\begin{equation}
	\mathcal{L}_\mathrm{cyc} = \lVert f^s_i - \epsilon^s_i \left(D_n \left(\left[\hat{a}^s_i, \hat{v}^s_i \right] \right) \right) \rVert_F^2,
\label{eq11}
\end{equation}
which encourages faithful recovery of both the raw input and its view-specific features. However, low-altitude views can dominate under the partial order, suppressing high-altitude signals. To further mitigate this and reduce feature ambiguity, we enforce that features of the same action across different views are more similar than features of different actions within the same view.


\subsection{Training and Inference}

After all modules, we aggregate the refined action features into a global representation:
\begin{equation}
	\hat{a}_{\mathrm{all}} = \sum_{i=1}^{n} c_i \hat{a}_i,
\end{equation}
which is used to predict the action category via the cross-entropy loss:
\begin{equation}
	\mathcal{L}_{\mathrm{ce}} = -\sum \hat{y} \log \left(\hat{a}_{\mathrm{all}}\right),
\end{equation}
where $\hat{y}$ is the one-hot ground truth.

The complete training objective combines three components:
\textit{1) Classification:} $\mathcal{L}_{\mathrm{ce}}$ for the common feature extractor.
\textit{2) Decoupling:} $\mathcal{L}_{\mathrm{dn}} = \mathcal{L}_{\mathrm{rec}} + \mathcal{L}_{\mathrm{cyc}}$ to separate action and view features.
\textit{3) Guidance:} $\mathcal{L}_{\mathrm{gn}} = \mathcal{L}_{\mathrm{kt}} + \mathcal{L}_{\mathrm{fd}} + \mathcal{L}_{\mathrm{vg}}$ for partial-order-based feature transfer and graph discretization.
The total loss is:
\begin{equation}
	\mathcal{L}_{\mathrm{all}} 
	= \gamma_{\mathrm{ce}} \mathcal{L}_{\mathrm{ce}}
	+ \gamma_{\mathrm{dn}} \mathcal{L}_{\mathrm{dn}}
	+ \gamma_{\mathrm{gn}} \mathcal{L}_{\mathrm{gn}},
\end{equation}
where $\gamma_{\mathrm{ce}}$, $\gamma_{\mathrm{dn}}$, and $\gamma_{\mathrm{gn}}$ balance each term. During inference, we compute $\hat{a}_{\mathrm{all}}$ as above and select the action label via $\arg\max(\hat{a}_{\mathrm{all}})$.


\section{Experimental Results}

\subsection{Datasets and Implementation Details}


\subsubsection{High-Altitude UAV View.} 

\textsc{Drone-Action}~\cite{perera2019drone} contains 240 high-altitude UAV videos covering 13 outdoor actions, recorded at 1920$\times$1080 resolution and 25 FPS. \textsc{MOD20}~\cite{PereraLOC20} provides a multi-view analysis of outdoor human actions with both aerial and ground footage across 2,324 videos at 720$\times$720 resolution and 29.97 FPS. \textsc{UAV}~\cite{LiLZNWL21} dataset includes 67,428 video sequences spanning 155 action classes from diverse aerial and vertical perspectives.


\subsubsection{Low-Altitude ground View.} 

\textsc{NTU-RGB+D}~\cite{ShahroudyLNW16} comprises 56,880 samples over 60 action classes, all recorded from ground-level cameras. It serves as a standard benchmark for multi-view ground action recognition.

\subsubsection{Settings and Metrics.} 

We evaluate under two settings: 
\textit{1) UAV View}, using only UAV datasets to assess multi-altitude representation; and 
\textit{2) Cooperative UAV and Ground Views}, augmenting with \textsc{NTU-RGB+D} to guide UAV performance. We follow standard protocols and report Top-1 accuracy for all experiments.


\subsubsection{Implementation Details.}

We adopt X3D~\cite{Feichtenhofer20} as our backbone, known for its efficiency in multi-view~\cite{LiuZZJWL23} and UAV-based~\cite{KothandaramanGW22} tasks. We insert a 2D temporal convolution (kernel size 3) after each 3D spatial convolution, followed by BN and ReLU. A dropout rate of 0.9 is applied before the final FC layer. We train with Adam~\cite{Kingma2014AdamAM}, an initial learning rate of 1e-3 decayed by 1e-5, for 300 epochs on four NVIDIA Tesla V100 GPUs (16 GB each).



	

\subsection{Comparison with State-of-the-Art Methods}

\cref{table1} compares POG-MVNet against several baselines~\cite{FayyazRDN0GG21, Cheng2022StochasticBA, LinGWH22} on \textsc{Drone-Action}, \textsc{MOD20}, and \textsc{UAV}. POG-MVNet consistently outperforms UniFormerV2~\cite{abs-2211-09552}, with accuracy gains of 16.5\%, 1.6\%, and 10.8\% on \textsc{Drone-Action}, \textsc{MOD20}, and \textsc{UAV}, respectively, demonstrating its adaptability to multi-view variations. 

Against ground-based multi-view methods, POG-MVNet also shows strong superiority: it improves over DRDN by 4.7\%, 1.5\%, and 3.0\% and over DVANer by 5.3\%, 1.7\%, and 3.7\% on \textsc{UAV}, \textsc{MOD20}, and \textsc{Drone-Action}, respectively. These results highlight its ability to model and integrate structured view relationships for more effective feature fusion. 

Moreover, POG-MVNet is computationally efficient, using only 15.2 M parameters, substantially fewer than DRDN (35.8 M) and ASAT (61.8 M), while maintaining robust accuracy, making it well-suited for resource-constrained UAV applications.

\begin{figure}[t]
	\centering
	\includegraphics[width = \linewidth]{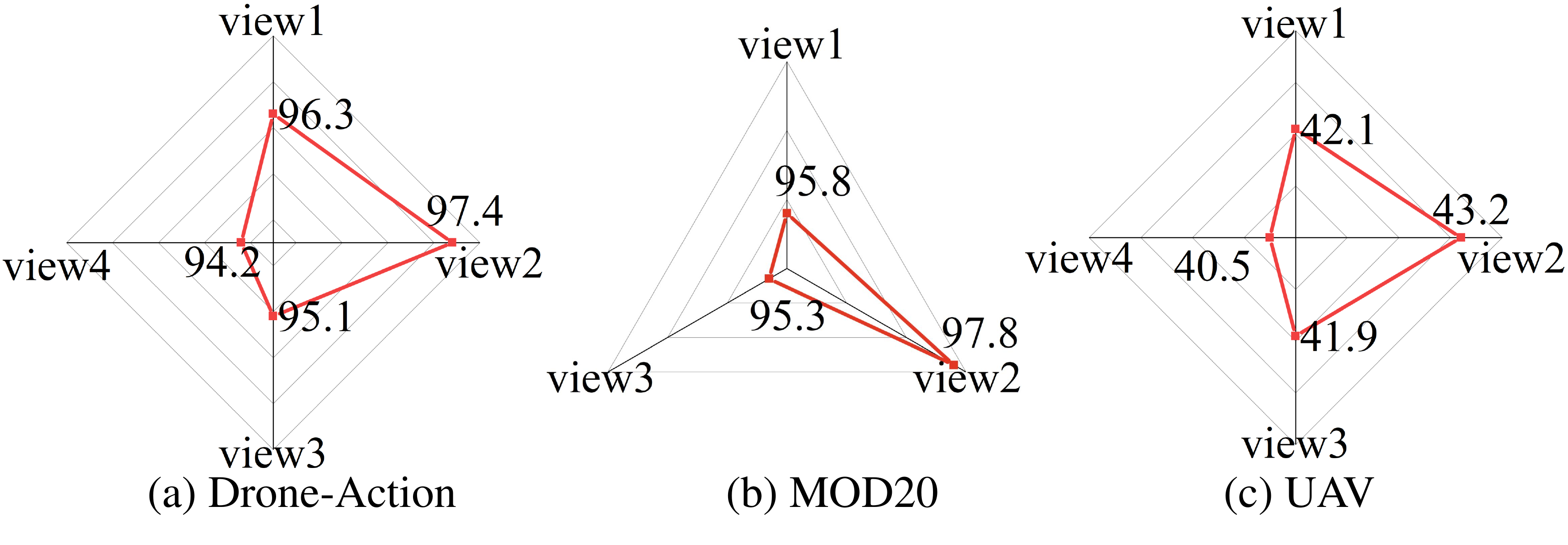} 
	\caption{\textbf{Performance comparison for different numbers of view partitions on \textsc{Drone-Action}, \textsc{MOD20}, and \textsc{UAV}.} \textsc{MOD20} is officially defined with four views, so no experiments with five views were conducted.}
	\label{fig-multi-views}
\end{figure}

\subsubsection{View Partition Analysis.}

We evaluate the impact of different view counts (2-5) on \textsc{Drone-Action}, as shown in \cref{fig-multi-views}. Three views yield the best performance: using only two views restricts cross-view information exchange and hampers the model’s ability to bridge altitude gaps, while using four or five views introduces redundant information that dilutes feature transfer and leads to similar action representations.

\begin{figure}[t]
	\centering
	\includegraphics[width = \linewidth]{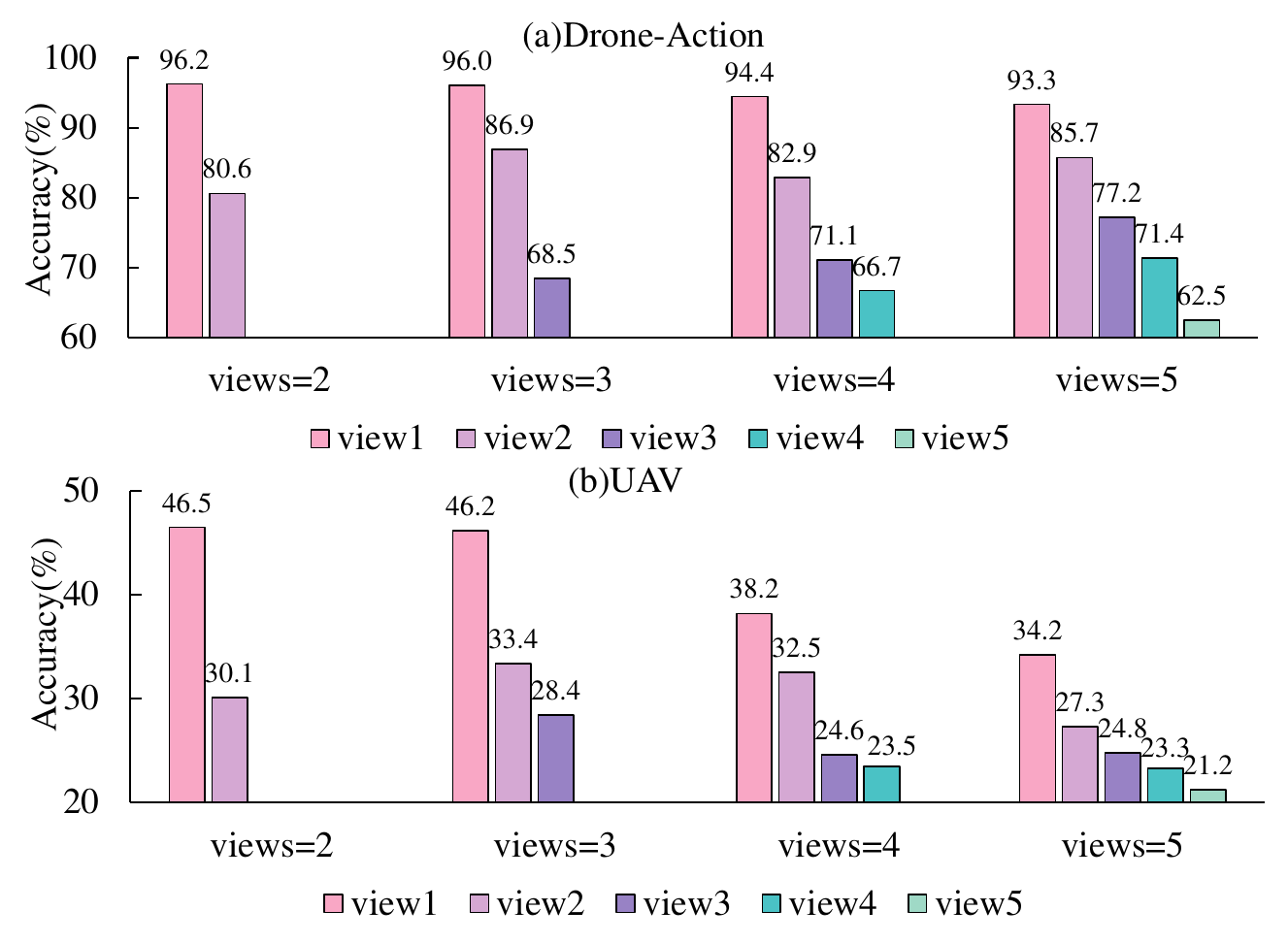} 
	\caption{\textbf{Performance comparison of each individual view under different numbers of view partitions on \textsc{Drone-Action} and \textsc{UAV}.} ``Views'' denotes the number of partitions.}
	\label{fig-multi-view-analysis}
\end{figure}

\subsubsection{Partial Order Relation Analysis.}


\cref{fig-multi-view-analysis} plots action recognition accuracy by view on \textsc{Drone-Action} and \textsc{UAV}, without using view labels. With two views, accuracies are 96.2\% (view 1) and 80.6\% (view 2). As we increase to five views, view 1 falls to 93.3\% and view 5 to 62.5\%. This widening gap, for example, view 3 at 68.5\% with three views, reveals a clear partial order among views, where lower-altitude views dominate and higher-altitude views lag. These results underscore the necessity of view-aware weighting in POG-MVNet to account for unequal contributions across views.

\begin{table}[t]
	\centering
	\setlength{\tabcolsep}{1mm}
	\small
	\begin{tabular}{c|l|l}
	\toprule[1.1pt]
	Views & Guide Strategy & Acc. (\%) \\
	\midrule
	\multirow{5}[4]{*}{2} 
	& None & 90.2 \\
	\cmidrule(lr){2-3}
	& $v_1 \to v_2$ & 94.2 ({$\uparrow$ 4.0}) \\
	& $v_2 \to v_1$ & 87.5 ({$\downarrow$ 2.7}) \\
	\cmidrule(lr){2-3}
	& \textsc{NTU-RGB+D} $\to$ \textsc{Drone-Action} & 95.8 ({$\uparrow$ 5.6}) \\
	& \textsc{Drone-Action} $\to$ \textsc{NTU-RGB+D} & 88.9 ({$\downarrow$ 1.3}) \\
	\midrule
	\multirow{9}[4]{*}{3}
	& None & 91.7 \\
	\cmidrule(lr){2-3}
	& $v_1 \to v_2$ & 95.8 ({$\uparrow$ 4.1}) \\
	& $v_1 \to v_3$ & 93.1 ({$\uparrow$ 1.4}) \\
	& $v_2 \to v_3$ & 94.5 ({$\uparrow$ 2.8}) \\
	& $v_1 \to v_2$ \& $v_2 \to v_3$ & 97.4 ({$\uparrow$ 5.7}) \\
	\cmidrule(lr){2-3}
	& $v_2 \to v_1$ & 90.2 ({$\downarrow$ 1.5}) \\
	& $v_3 \to v_1$ & 87.5 ({$\downarrow$ 4.2}) \\
	& $v_3 \to v_2$ & 90.2 ({$\downarrow$ 1.5}) \\
	& $v_3 \to v_2$ \& $v_2 \to v_1$ & 88.9 ({$\downarrow$ 2.8}) \\
	\bottomrule[1.1pt]
	\end{tabular}
	\caption{\textbf{Top-1 accuracy (\%) of different guide strategies in POG-MVNet on \textsc{Drone-Action} and \textsc{NTU-RGB+D}.} $v_i$ denotes the $i$-th view.}
	\label{table-relation}
\end{table}

\subsubsection{Guide Strategy Analysis.}



\cref{table-relation} reports multi-view feature-transfer results on \textsc{Drone-Action} for views = 2 and 3. For views = 2, transferring from $v_{1}\to v_{2}$ yields a 4.0\% improvement over average fusion, whereas $v_{2}\to v_{1}$ reduces accuracy by 2.7\%, suggesting that high-altitude views introduce noise and degrade low-altitude representations, bottom-up transfer is thus more effective than top-down guidance. In a cross-dataset experiment using five shared actions from \textsc{NTU-RGB+D}, POG-MVNet gains 5.6\% when transferring from ground (low) to UAV views but loses 1.3\% in the reverse direction, further confirming that low-altitude features are more transferable and generalizable. For views = 3, direct transfers from $v_{1}\to v_{2}$ outperform both $v_{1}\to v_{3}$ and $v_{2}\to v_{3}$, likely due to closer feature similarity between low- and mid-altitude perspectives, and the sequential transfer $v_{1}\to v_{2}\to v_{3}$ achieves the best result, in line with the partial-order view hierarchy. All reverse sequences degrade performance (\textit{e.g.}, $v_{2}\to v_{1}$ drops by 1.5\%), underscoring the value of bottom-up knowledge transfer.

\begin{table}
	\centering
	\small
	\begin{tabular}{ccc|ccc}
	\toprule[1.1pt]
	Baseline & VDG & APOG & \textsc{Drone} & \textsc{MOD20} & \textsc{UAV} \\
	\midrule
	\CIRCLE & \Circle & \Circle & 83.4 & 85.2 & 32.3 \\
	\CIRCLE & \CIRCLE & \Circle & 91.7 & 90.1 & 38.4 \\
	\CIRCLE & \Circle & \CIRCLE & 93.1 & 92.5 & 39.6 \\
	\CIRCLE & \CIRCLE & \CIRCLE & \textbf{97.4} & \textbf{97.8} & \textbf{43.2} \\
	\bottomrule[1.1pt]
	\end{tabular}
	\caption{\textbf{Comparison of Top-1 accuracy (\%) for different POG-MVNet variants on \textsc{Drone-Action}, \textsc{MOD20}, and \textsc{UAV}.} Best results are shown in bold.}
	\label{table-module}
\end{table}

\subsection{Ablation Studies}


\cref{table-module} presents module-wise ablations. Both the APOG and VDG units improve upon the X3D baseline~\cite{Feichtenhofer20}, with APOG yielding the largest gain by enabling effective feature transfer from low- to high-altitude views. Combining both modules achieves the highest overall accuracy, demonstrating their complementary benefits. (\underline{More details can be found in the supplementary materials.})

\begin{table}[t] 
	\centering
	\small
	\setlength{\tabcolsep}{1mm}
	\begin{tabular}{c|l|ccc} 
	\toprule[1.1pt]
	\multicolumn{2}{c|}{Loss Weight} & \textsc{Drone} & \textsc{MOD20} & \textsc{UAV} \\
	\midrule
	\multirow{4}{*}{\makecell{\underline{$\gamma_\mathrm{ce}$}, $\gamma_\mathrm{dn}$, $\gamma_\mathrm{gn}$ \\ ($\gamma_\mathrm{dn}$ = 1, $\gamma_\mathrm{gn}$ = 1)}}
	& $\gamma_\mathrm{ce}$ = 0.01 & 90.2 & 91.2 & 34.5 \\
	& $\gamma_\mathrm{ce}$ = 0.1 & 91.7 & 93.2 & 36.3 \\
	& $\gamma_\mathrm{ce}$ = 1 & {95.8} & {97.2} & {42.5} \\
	& $\gamma_\mathrm{ce}$ = 10 & 94.4 & 94.7 & 39.1 \\
	\midrule
	\multirow{4}{*}{\makecell{$\gamma_\mathrm{ce}$, \underline{$\gamma_\mathrm{dn}$}, $\gamma_\mathrm{gn}$ \\ ($\gamma_\mathrm{ce}$ = 1, $\gamma_\mathrm{gn}$ = 1)}}
	& $\gamma_\mathrm{dn}$ = 0.01 & 91.7 & 91.4 & 35.1 \\
	& $\gamma_\mathrm{dn}$ = 0.1 & 93.1 & 94.0 & 38.9 \\
	& $\gamma_\mathrm{dn}$ = 1 & {95.8} & {97.2} & {42.5} \\
	& $\gamma_\mathrm{dn}$ = 10 & 93.1 & 95.3 & 39.2 \\
	\midrule
	\multirow{4}{*}{\makecell{$\gamma_\mathrm{ce}$, $\gamma_\mathrm{dn}$, \underline{$\gamma_\mathrm{gn}$} \\ ($\gamma_\mathrm{ce}$ = 1, $\gamma_\mathrm{dn}$ = 1)}}
	& $\gamma_\mathrm{gn}$ = 0.01 & 94.4 & 95.1 & 41.2 \\
	& $\gamma_\mathrm{gn}$ = 0.1 & \textbf{97.4} & \textbf{97.8} & \textbf{43.2} \\
	& $\gamma_\mathrm{gn}$ = 1 & 95.8& 97.2 & 42.5 \\
	& $\gamma_\mathrm{gn}$ = 10 & 91.7 & 92.6 & 36.9 \\
 	\midrule
	\multirow{4}{*}{\makecell{\underline{$\gamma_\mathrm{ce}$}, $\gamma_\mathrm{dn}$, $\gamma_\mathrm{gn}$ \\ ($\gamma_\mathrm{dn}$ = 1, $\gamma_\mathrm{gn}$ = 0.1)}}
	& $\gamma_\mathrm{ce}$ = 0.01 & 91.7 & 92.3 & 36.5 \\
	& $\gamma_\mathrm{ce}$ = 0.1 & 93.1 & 94.6 & 39.1 \\
	& $\gamma_\mathrm{ce}$ = 1 & \textbf{97.4} & \textbf{97.8} & \textbf{43.2} \\
	& $\gamma_\mathrm{ce}$ = 10 & 94.4 & 95.7 & 39.8 \\
 \midrule
	\multirow{4}{*}{\makecell{$\gamma_\mathrm{ce}$, \underline{$\gamma_\mathrm{dn}$}, $\gamma_\mathrm{gn}$ \\ ($\gamma_\mathrm{ce}$ = 1, $\gamma_\mathrm{gn}$ = 0.1)}}
	& $\gamma_\mathrm{dn}$ = 0.01 & 93.1 & 92.5 & 37.2 \\
	& $\gamma_\mathrm{dn}$ = 0.1 & 94.4 & 95.1 & 39.5 \\
	& $\gamma_\mathrm{dn}$ = 1 & \textbf{97.4} & \textbf{97.8} & \textbf{43.2} \\
	& $\gamma_\mathrm{dn}$ = 10 & 93.1 & 95.9 & 40.1 \\
	\bottomrule[1.1pt]
	\end{tabular}
 	\caption{\textbf{Comparison of Top-1 accuracy (\%) on \textsc{Drone-Action}, \textsc{MOD20}, and \textsc{UAV} for different loss weights.} Best results are shown in bold.}
	\label{table-loss}
\end{table} 

\subsubsection{Loss Parameter Analysis.} 

We conduct a detailed loss analysis for POG-MVNet on \textsc{Drone-Action}, \textsc{MOD20}, and \textsc{UAV}, as presented in \cref{table-loss}. The total loss function, $\mathcal{L}_{\mathrm{all}}$, integrates three components: classification loss $\mathcal{L}_{\mathrm{ce}}$, decoupled loss $\mathcal{L}_{\mathrm{dn}}$, and guide loss $\mathcal{L}_{\mathrm{gn}}$. Following the Gibbs-sampling-inspired iterative scheme of~\cite{HarrisonMARATLJ20}, we adjust each weight independently, holding the others constant, over two rounds to identify their optimal values. We find that setting $\gamma_{\mathrm{ce}}=1$, $\gamma_{\mathrm{dn}}=1$, and $\gamma_{\mathrm{gn}}=0.1$ yields the best performance: 95.8\%, 97.2\%, and 42.5\% peak accuracies on \textsc{Drone-Action}, \textsc{MOD20}, and \textsc{UAV}, respectively. Lowering either $\gamma_{\mathrm{dn}}$ or $\gamma_{\mathrm{gn}}$ leads to noticeable drops in accuracy, whereas modest increases in these weights improve results, most notably on \textsc{UAV}. These findings underscore the necessity of carefully balancing the three loss terms to maximize feature learning and recognition performance.

\begin{figure}[t]
	\centering
	\includegraphics[width = 0.9\linewidth]{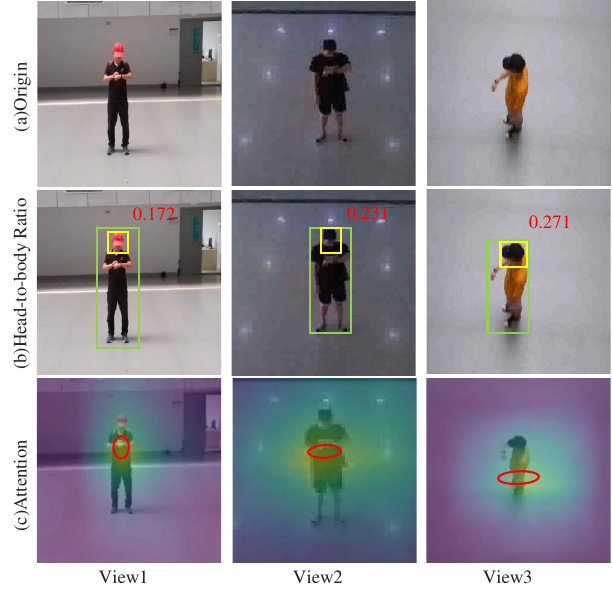}
	\caption{\textbf{Visualization of head-to-body ratio and action feature attention on \textsc{UAV}.} (b) Green and yellow boxes denote body and head detections, respectively; red numbers show the head-to-body ratio (smaller values correspond to lower-altitude views). (c) Red circles mark the centers of attention for the action.}
	\label{fig-action}
\end{figure}


\subsubsection{Visualization.}

\cref{fig-action} visualizes attention maps on \textsc{UAV} for the action ``Look at the watch'', revealing stark differences across views. In the low-altitude view (view 1), attention concentrates sharply on the hand, the key region, thanks to proximity, which accentuates fine details like hand movements. In contrast, the high-altitude view (view 3) produces a more diffuse attention distribution due to the broader field of view and greater distance, which obscures action-specific features. This comparison highlights the critical role of low-altitude views in capturing precise action cues and underscores how viewing altitude can significantly influence recognition accuracy.


\section{Conclusion}

In this paper, we identify a previously overlooked partial order structure among UAV views, where action recognition accuracy consistently degrades with increasing altitude due to greater visual ambiguity and reduced motion cues. To address this challenge, we propose the Partial Order Guided Multi-View Network (POG-MVNet), which explicitly models the hierarchical nature of UAV views to enhance cross-view action recognition. Our framework comprises a View Partition (VP) module that segments views using the head-to-body ratio, an Order-aware Feature Decoupling (OFD) module that disentangles action- and view-specific features under partial order guidance, and an Action Partial Order Guide (APOG) unit that enables progressive knowledge transfer from easier to harder views. Extensive experiments on multiple UAV benchmarks and ground multi-view datasets confirm that our approach significantly outperforms conventional single-view and multi-view baselines.






\bibliography{aaai2026}



\end{document}